\documentclass[pmlr]{jmlr}

\RequirePackage{graphicx}

\usepackage[utf8]{inputenc}
\usepackage[T1]{fontenc}
\usepackage{lmodern}
\usepackage{microtype}
\usepackage{hyperref}
\usepackage{url}
\usepackage{doi}
\usepackage[x11names]{xcolor}
\usepackage{textcomp}
\usepackage{manyfoot}
\usepackage{listings}
\usepackage[bottom]{footmisc}
\usepackage{authblk}

\usepackage{graphicx}
\usepackage{booktabs}
\usepackage{longtable}
\usepackage{array}
\usepackage{multirow}

\usepackage{mathtools}
\usepackage{amsmath}
\usepackage{amsfonts}
\usepackage{amssymb}
\usepackage{centernot}
\usepackage{nicefrac}

\PassOptionsToPackage{hyphens}{url}

\makeatletter                             %
\def\set@curr@file#1{\def\@curr@file{#1}} %
\makeatother                              %
\usepackage[load-configurations=version-1]{siunitx}

\usepackage{cleveref}
\crefname{section}{Section}{Sections}
\crefname{figure}{Figure}{Figures}
\crefname{table}{Table}{Tables}

\jmlrvolume{252}
\jmlryear{2024}
\jmlrworkshop{Machine Learning for Healthcare}

\title{MedTsLLM: Leveraging LLMs for Multimodal Medical Time Series Analysis}

\author[*1]{Nimeesha Chan}
\author[*1]{Felix Parker}
\author[2]{William Bennett}
\author[3]{Tianyi Wu}
\author[4]{Mung Yao Jia}
\author[2]{James Fackler}
\author[1]{Kimia Ghobadi}

\affil[1]{Department of Civil and Systems Engineering, Johns Hopkins University}
\affil[2]{Department of Anesthesiology and Critical Care Medicine, Johns Hopkins University}
\affil[3]{Department of Applied Mathematics and Statistics, Johns Hopkins University}
\affil[4]{Department of Computer Science, Johns Hopkins University}

\hypersetup{
    pdftitle={Leveraging LLMs for Multimodal Medical Time Series Analysis},
	pdfsubject={},
	pdfauthor={Nimeesha Chan, Felix Parker, William Bennett, Tianyi Wu, Mung Yao Jia, James Fackler, Kimia Ghobadi},
	pdfkeywords={},
}

\begin{document}
\maketitle
\footnotetext{\textsuperscript{*}These authors contributed equally to this work.}

\vspace{-2em}
\begin{abstract}
The complexity and heterogeneity of data in many real-world applications pose significant challenges for traditional machine learning and signal processing techniques. For instance, in medicine, effective analysis of diverse physiological signals is crucial for patient monitoring and clinical decision-making and yet highly challenging. 
We introduce MedTsLLM, a general multimodal large language model (LLM) framework that effectively integrates time series data and rich contextual information in the form of text to analyze physiological signals, performing three tasks with clinical relevance: semantic segmentation, boundary detection, and anomaly detection in time series. These critical tasks enable deeper analysis of physiological signals and can provide actionable insights for clinicians. We utilize a reprogramming layer to align embeddings of time series patches with a pretrained LLM's embedding space and make effective use of raw time series, in conjunction with textual context. Given the multivariate nature of medical datasets, we develop methods to handle multiple covariates. We additionally tailor the text prompt to include patient-specific information. Our model outperforms state-of-the-art baselines, including deep learning models, other LLMs, and clinical methods across multiple medical domains, specifically electrocardiograms and respiratory waveforms. MedTsLLM presents a promising step towards harnessing the power of LLMs for medical time series analysis that can elevate data-driven tools for clinicians and improve patient outcomes.\footnotemark[1]
\end{abstract}

\footnotetext[1]{Code is available here: \href{https://github.com/flixpar/med-ts-llm}{https://github.com/flixpar/med-ts-llm}}

\section{Introduction}
\label{sec:intro}
Precision medicine and personalized decision support tools have long aimed to leverage multimodal patient data, from free-form text notes to semi-structured electronic health records (EHR) to high-frequency physiological signals, to better capture the complex, high-dimensional patient state and the provider responses. However, combining these heterogeneous data types has been challenging, with early approaches building bespoke models for single data types and tasks. While the advent of transformer architectures enabled deeper insight from merging modalities, it also required meticulous feature engineering and alignment.

We propose utilizing the knowledge and higher-level reasoning that large language models (LLMs) acquire during pretraining to interpret multidimensional, high-frequency physiological signals and produce high-fidelity output. LLMs, trained on vast datasets and adaptable to various downstream tasks, have ushered in a new era of multimodal foundation models \citep{yin2023survey}. While quick to be adopted in other domains, healthcare has lagged behind, partly due to a lack of high-quality labeled datasets. Current medical LLMs mostly focus on image-text pairs \citep{ghosh2024clipsyntel}, EHRs \citep{li2024scoping}, or clinical notes \citep{jung2024enhancing}. Only recently have LLMs been explored with physiologic signals, and usually for classification \citep{liu2024zero} or report generation \citep{wan2024electrocardiogram}.

The clinical motivation for this work is to leverage the power of large language models to find hidden patterns in time series data. Certainly, single-dimensional time series, such as temperature graphs and pulse oximeter waveforms are comfortably analyzed. However, these ``simple” single-dimensional time series can only be analyzed by clinicians as instantaneous ``snapshots” and longitudinal patterns are lost (e.g., rates of change or even simply, counts over time of like events). It becomes increasingly challenging for clinicians to extract 
nuanced, meaningful patterns in time series data when the data is overwhelmingly large and patterns are necessarily multi-dimensional, for instance, cardio-pulmonary interactions that can only be detected with simultaneous analysis of blood pressure waveforms, ventilator waveforms, electrocardiogram (ECG), and even the lower fidelity time series dimensions available in laboratory data.
Harnessing the depth of pattern recognition offered by large language models can pave the way to a deeper understanding of clinically significant patterns across multi-dimensional physiological data.

By aligning multivariate time series with patient context using a patch reprogramming layer, our unified framework performs clinically useful tasks like semantic segmentation, boundary detection, and anomaly detection. At a high level, boundary detection splits signals into periods like breaths or beats. Semantic segmentation further splits time series into distinct, meaningful segments.
Anomaly detection identifies periods within the signals that deviate from normal. These tasks are critical for interpreting waveforms, such as those from ECGs, breathing, and other vital signals, and inherently require domain-specific knowledge for phase identification, rendering a one-size-fits-all method impractical, and highlighting the need for integrating unstructured data. 
Compounded to that, medical signals often contain interactions between different systems (e.g. cardio-pulmonary interactions) that need to be considered in conjunction with each other. To better handle these covariates, we redesign the structure of the data at various points in our architecture, propose multiple covariate-handling methods, and investigate their tradeoffs in performance and accuracy.

We primarily focus on mechanical ventilators to segment breath phases to provide insights into respiratory mechanics, which can guide decisions on weaning or escalating care for these patients.
We further test and validate our models on a set of publicly available medical datasets to ensure our framework generalizes across well-studied ECG domains, as well as lesser-studied respiratory waveforms. Our model outperforms state-of-the-art methods in segmenting ECGs and ventilator waveforms, and in detecting arrhythmias. Such capabilities enable downstream applications that could transform critical care medicine, e.g., identifying complex ECG patterns for early diagnosis of life-threatening conditions, or accurate ECG segmentation to measure prognostic parameters such as heart rate variability and QT interval. 
The multimodal nature of MedTsLLM, along with explicit considerations of covariates, provides an opportunity for further use of medical data to gain more comprehensive insights. In this paper, our main contributions include:
\begin{itemize} \setlength{\itemsep}{0pt}
    \item We propose a framework that uses the power of a pretrained LLM %
    to harness learnings from unstructured and semi-structured data through natural language, with high-dimensional time series signals.
    \item We design novel methods to more holistically capture the strong correlations between covariates in time series, and discuss their applicability in various settings.
    \item We construct prompts that provide the LLM with patient-specific information, alongside dataset description, task instructions and sample statistics--all of which contextualizes the provided time series.
    \item We introduce three novel time series tasks in the space of using text-time series fusion LLMs: boundary detection, semantic segmentation, and anomaly detection.
    \item Our model outperforms state-of-the-art baselines, including transformer-based models, LLM-based models, traditional time series analysis models, and domain-specific methods, across multiple medical and non-medical tasks and datasets.
\end{itemize}

\subsection*{Generalizable Insights about Machine Learning in the Context of Healthcare}
The integration of advanced computational methods, such as large language models (LLMs), with medical time series analysis holds immense promise for advancing data-driven support tools for clinical decision-making and ultimately improving patient outcomes. The main advantages of employing LLMs for medical tasks are to leverage the extensive knowledge and reasoning abilities of LLMs and to capitalize on the potential of multimodal data in healthcare. Our model achieves superior performance on both typical time series analysis tasks like anomaly detection and more specialized, clinically insightful tasks like segmentation across multiple medical applications and datasets. This generalizability of our framework to several tasks and medical applications is promising evidence that LLMs may be useful for diverse tasks in healthcare.

Our ablation studies demonstrate that our LLM-based framework effectively utilizes both time series data and text information. Notably, we show that including patient-specific information in the text prompt improves model performance. Thus, LLMs enable the use of natural language prompts as a medium to combine both unstructured and semi-structured information, as is common in EHRs. %
We develop a range of techniques to better capture the relationships between multivariate physiological signals, and our findings indicate that how covariates are handled significantly impacts model performance. In conclusion, our work leverages the power of LLMs and their ability to integrate diverse data types, which will open up new avenues for more comprehensively consolidating patient information to perform clinically useful downstream tasks, ultimately leading to more accurate diagnoses, targeted treatments, and improved patient care.

\section{Related Work}
\label{sec:relatedwork}
There is a rich and fast-growing literature on the use of large language models and on the use of data in healthcare tasks. We focus on the two most relevant fields of using LLMs for time series analysis, both general and medical time series, and methods used in healthcare tasks of interest, namely, semantic segmentation, boundary detection, and anomaly detection.

\paragraph{Time Series Prediction Tasks.}

The three tasks we focus on in this work are semantic segmentation, boundary detection, and anomaly detection. %
Semantic segmentation entails partitioning an input into contiguous segments that represent an object or event, and classifying each segment. This problem has been studied extensively in the context of computer vision, in which it is natural to segment objects in an image and classify them \citep{hao2020brief}.
It is also a critical task in time series analysis, where the aim is to segment distinct events or phases \citep{keogh2004segmenting}, although it sometimes referred to as just segmentation.
Semantic segmentation is typically formulated as a point-wise classification problem, in which each point is classified independently by a trained classification model. Many such point-wise classifiers have been developed, with recent efforts focusing on deep learning models \citep{NIPS2019_8692,Gaugel_2023}.
Boundary prediction is a closely related task that similarly involves partitioning a signal into discrete segments, but in which we do not have semantic labels for each segment.
In the medical domain, these tasks have been studied for ECG and breath waveform phase detection, utilizing a range of statistical \citep{4122029, 10.1093/chemse/bjy045} and machine learning \citep{Moskalenko_2019,londhe2021semantic,liang2022ecg_segnet} approaches.
Three notable related works perform breath segmentation \citep{chong2021, 10.1093/chemse/bjy045, 10.1007/978-3-031-51485-2_29}, but all require a certain set of waveforms as input, which are not always available.

General unsupervised time series anomaly detection has been extensively studied in the literature and there are many diverse approaches to the problem \citep{lu2008network,salem2014anomaly,pena2013anomaly}.
Recently, deep learning based methods, and in particular transformer-based architectures, have garnered significant attention in this problem setting due to their flexibility and strong performance \citep{wu2022timesnet,xu2018unsupervised,gao2020robusttad}.
While these methods have had great success on a variety of benchmark datasets, they largely focus exclusively on time series signals without additional unstructured context, which can be of great importance, particularly in specialized domains such as ECG analysis.
Many domain-specific methods have also been investigated for time series anomaly detection in medical settings that overlap with the datasets we focus on in this work. In particular, various methods that target anomaly detection in ECG signals have been proposed \citep{li2020survey,sivapalan2022annet,alamr2023unsupervised}.
However, these methods tend to be tailored to the specific problem using knowledge of the data and types of anomalies, and thus cannot be readily adapted to other domains.

\paragraph{LLMs for Time Series Analysis.} 
LLMs have been applied to general time series tasks such as forecasting and classification by integrating time series data through prompt augmentation~\citep{xue2023promptcast, gruver2024large}, or utilizing pretrained backbones for downstream tasks~\citep{zhou2024one}. \citet{liu2023unitime} uses explicit domain identification information to allow forecasting strategy adaptation. Our study builds upon the work of \citet{jin2023time}  which introduces a ``reprogramming'' layer to project time series patches onto a pretrained LLM's embedding space, enabling it to make effective use of the raw time series in conjunction with textual context for forecasting. Our study adapts this work for the medical domain in addition to the following methodological contributions: (1) extending it to solve a set of time series tasks that are clinically relevant, (2) improving the way covariates are utilized, and (3) augmenting the text prompt to include patient-specific clinical information.

In the medical domain, LLMs have been used to analyze biomedical signals, particularly in electroencephalogram (EEG) and electrocardiogram (ECG) analyses, for tasks like automated report generation~\citep{duan2024dewave} and zero-shot disease detection~\citep{wang2024enhancing}. 
Multimodal LLM frameworks have been used to enhance disease risk quantification \citep{belyaeva2023multimodal} and pattern recognition using wearable data~\citep{liu2023large}. As far as we are aware, our method is the first to apply LLMs to the medical domain for time series tasks. In addition, our method more effectively utilizes time series input data with the aforementioned ``reprogramming'' layer and our adaptations.

\section{Methods}
\label{sec:methods}

\begin{figure}[htb]
\vspace{-.1cm}
    \centering
    \includegraphics[width=1\textwidth]{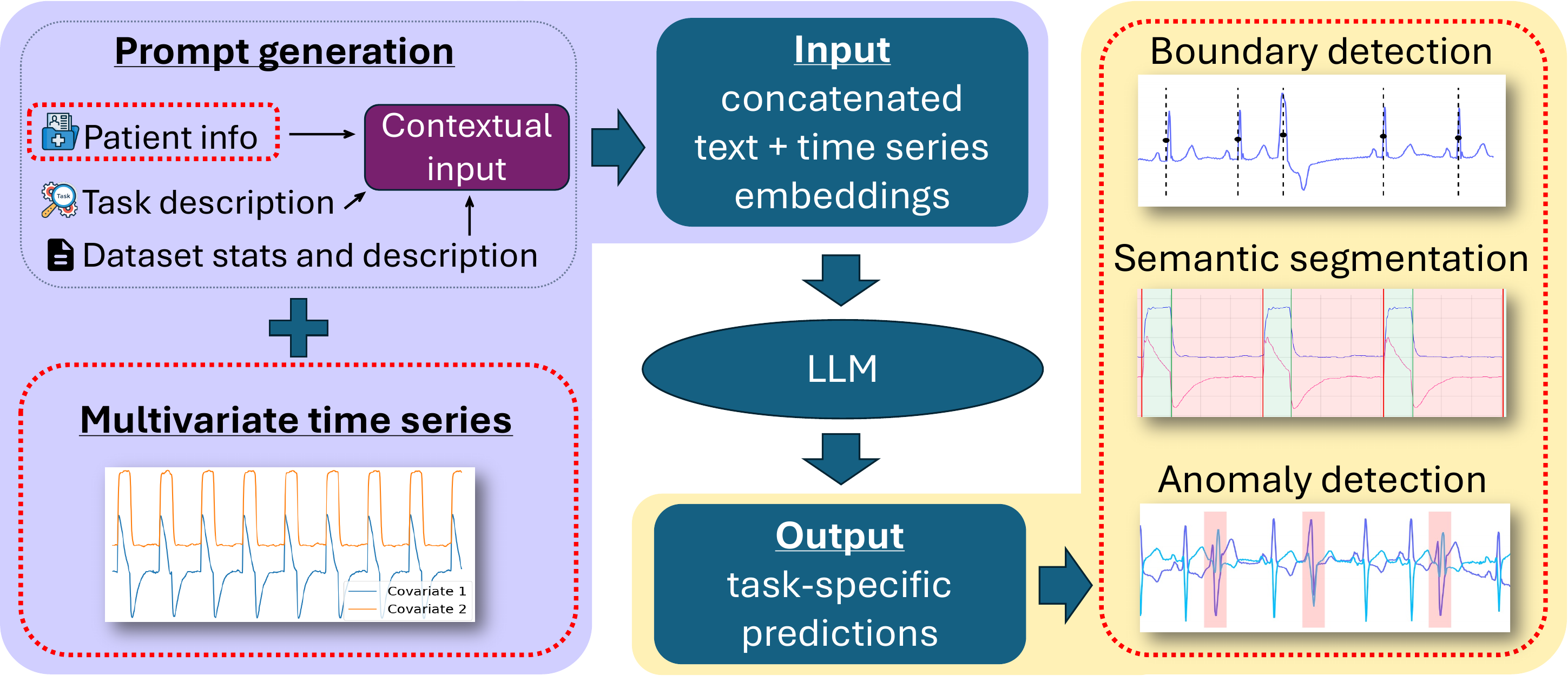}
    \vspace{-.8cm}
    \caption{\small In our proposed framework, the multimodal input consists of contextual input and raw time series data, which are both converted to embeddings. The concatenated embeddings are fed into a pretrained LLM. Output LLM embeddings are then used by task-specific methods to generate predictions. Our contributions are highlighted with red dotted lines.}
    \label{fig:overview}
    \vspace{-.4cm}
\end{figure}

\begin{figure}[htb]%
    \centering
    \includegraphics[width=1.0\textwidth, trim=15 10 10 10, clip=true]{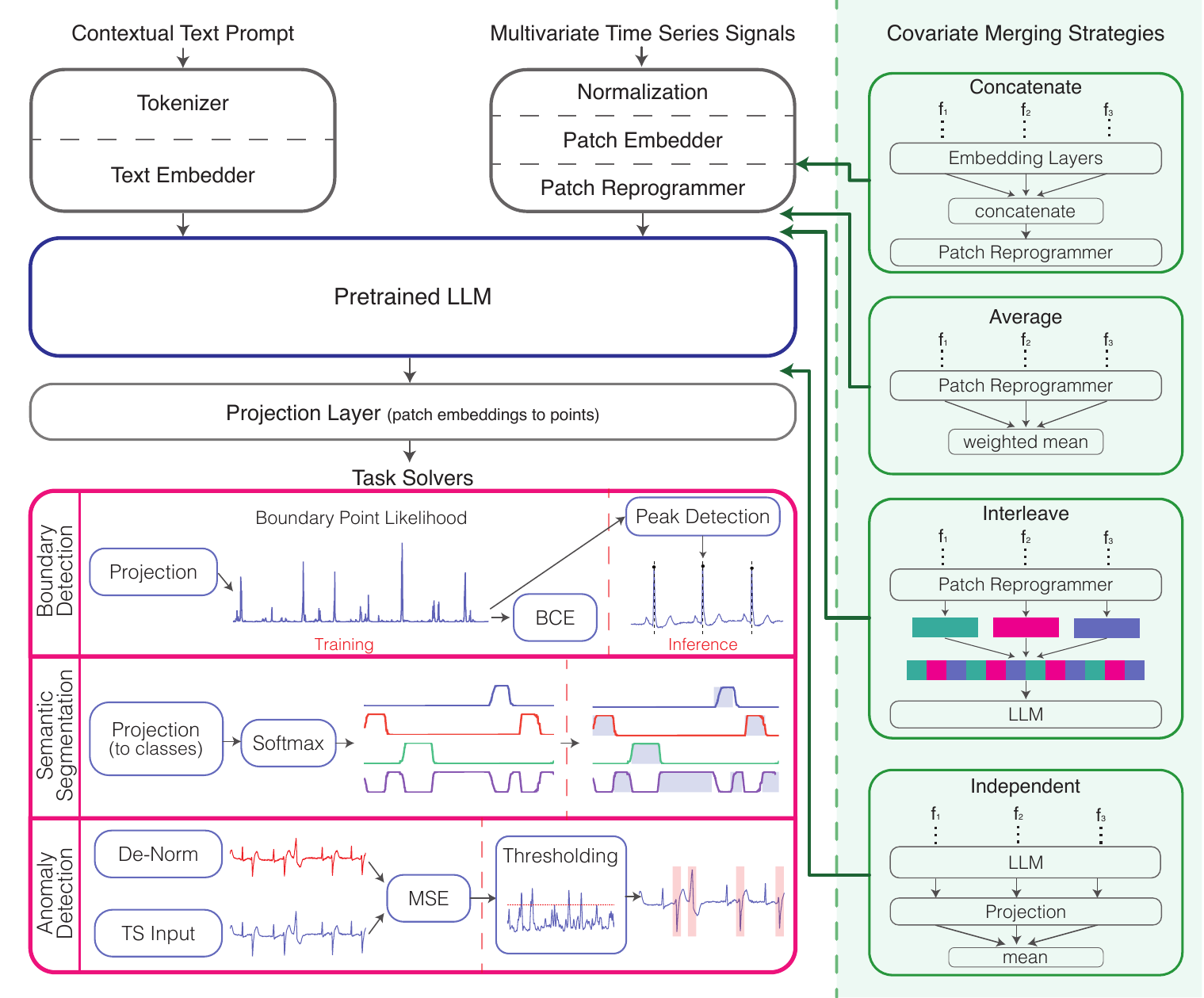}
    \vspace{-.3cm}
    \caption{\footnotesize A schematic overview of our proposed methodology. Time series and corresponding textual context are embedded separately and concatenated. While text passes through standard tokenization and embedding, time series is patched and transformed into embeddings, which are aligned to text embeddings using a patch reprogrammer layer. Covariate information is merged using one of our proposed strategies, and embeddings are fed into the LLM, and through a projection layer to produce raw predictions. These predictions are transformed to solve one of our selected analysis tasks using task-specific layers and processing. The covariate strategies (green) and task solvers (pink) are our primary methodological contributions. All plot examples in the task solvers section are real outputs of our model on the datasets used in this work. }
    \label{fig:model}
    \vspace{-.9cm}
\end{figure}

In this study, we develop a multimodal model that can adapt pretrained LLMs for multivariate time series task-solvers. The distinctive feature of this approach is its ability to process both raw time series data and natural language inputs. Our method consists of four core components: (1) prompt generation (\cref{sec:methods:prompt}), (2) time series embedding (\cref{sec:methods:embedding}), (3) a pretrained LLM (\cref{sec:methods:llm}), and (4) time series task-solvers (\cref{sec:methods:tasks}), as illustrated in \cref{fig:overview}, with more details provided in \cref{fig:model}.  
We use dataset, task, patient-specific, and time series information to construct a prompt with relevant context that instructs the LLM to solve the desired task. We split time series into patches and align the patch embeddings with text embeddings from the LLM so that it can utilize them effectively. For alignment, we adopt the patch reprogrammer introduced in \citet{jin2023time}, but extend it to incorporate covariates in the framework. We then feed the word and time series embeddings into state-of-the-art LLMs to analyze the text and time series together. Finally, we take the output embeddings of the LLM and use task-specific projection layers and processing to solve the selected time series analysis task.

The three tasks we focus on in this work are anomaly detection, boundary detection, and semantic segmentation.
Anomaly detection involves determining which periods of the input deviate from normal behavior.
Boundary detection requires finding points that split a time series up into periods that represent a single type of event, such as a heartbeat or breath.
Semantic segmentation is similar to boundary detection, but also determines a classification for each sequence, which is useful when there are multiple distinct types of events (e.g., ventilator-delivered breaths vs spontaneous breathing) or distinct event phases (e.g., inspiration and expiration) that are important to capture. These methods are further described in \cref{sec:methods:tasks}.

\subsection{Prompt Generation}
\label{sec:methods:prompt}

We aim to improve the performance of our task solvers by including relevant information that is not captured in the time series or other structured data such as clinical notes, diagnostic reports, and medical history, by capturing it as unstructured natural language and using it to inform time series predictions.
Therefore, it is critical that we provide the LLM with useful context and an effective prompt to maximize the power of its language modeling abilities. %
There are four main components of the context that we provide the LLM: dataset description, task description, a summary of dataset statistics and detailed patient-specific information  (see \Cref{fig:overview}). 
To leverage the LLM's pretraining, we encode all static information as text, rather than developing separate encoding mechanisms per data modality.

The dataset description provides the LLM with the relevant background information to explain the domain and context. It informs the LLM of what the problem setting is, what modality of data each feature represents, and how the data was collected. This information allows the model to apply any knowledge about the task and data that it may have learned in the pretraining phase. Specific prompts for each dataset can be found in \cref{tab:dataset-prompts}. Task instructions, important for instruction-tuned LLMs, guide the model’s expected output (e.g., “Identify the change points in the past 256 steps of data to segment the sequence.”). Some additional summary statistics (e.g. minimum and overall signal trend) are included for low-frequency complementary signals (less than 1 Hz) and encoded as text. 

Further information about the specific patient that the time series represents can be highly relevant to the predictions. Large amounts of medical data are stored as unstructured text, including clinician notes and other reports, which may provide important context about conditions or diagnoses that affect the time series inputs and outputs. The patient-specific information included in the prompt depends on what was available for each dataset, but generally contains patient demographics, administered medications, diagnoses, clinical notes, and additional medical history. To standardize the formatting, particularly for structured demographic information, patient-level data was encoded as JSON, which also improved the model's performance.

The dataset description, patient-level information, summary statistics, and task instruction were concatenated (in that order) into one string and sent through the pretrained LLM's tokenizer and embedding layer to produce token embeddings.
These are then concatenated with the time series patch embeddings, which are discussed in \cref{sec:methods:embedding}, to form the input to the LLM. We also evaluated the contribution of each component in \cref{tab:results:abl:prompt}.

\subsection{Time Series Encoding}
\label{sec:methods:embedding}

The main barrier to effectively using pretrained LLMs for time series analysis is encoding time series data in such a way that a pretrained LLM can make good use of it.
Tokenization of time series data encoded as text maps 1-3 digits at a time to a discrete embedding, which requires either very low numerical precision or splitting numbers into multiple tokens. This inefficient method results in suboptimal encodings. Our proposed patch embedding method overcomes these limitations by using multiple data points (typically 16 per patch) at full precision, which are then transformed into continuous embeddings.
We train a time series encoder for a pretrained, frozen LLM, to learn embeddings of patches of raw time series signals. Thus, the model can utilize multimodal text and time series inputs without converting signals to text, and without computationally expensive fine-tuning of the LLM.

Patching the input signals allows us to construct tokens with semantic resolution more similar to that of natural language.
We first take the input signal $x_{1:T} \in \mathbb{R}^T$ and normalize it using Reversible Instance Normalization (RevIN) \citep{kim2021reversible} to account for distribution shift over time.
From the normalized signal we construct a set of patch vectors $\{ x_{t:t+l-1} : i \in P, t = i s + 1 \}$, where $P = \{ i \in \mathbb{Z} : 0 \leq i \leq \frac{T-l}{s} \}$ is the set of patch indices, $l$ is the patch length, and $s$ is the stride. Each patch is then transformed into an embedding using a linear projection layer.

However, in order for a pretrained LLM to make effective use of these embeddings without extensive fine-tuning, the embeddings must be aligned with the distribution of pretrained token embeddings. It is a well understood problem in ML that models often perform poorly on a target domain if there is a significant gap between the target data distribution and the training distribution \citep{ben2010theory}. Therefore, the LLM backbone will likely not perform well in our model if there is a significant gap between the distribution of text embeddings and time series embeddings. This problem is recognized as a core challenge in multimodal ML \citep{gao2020survey}, and has received significant attention recently for vision-language models.

To address this, we utilize the patch reprogramming layer introduced in \citet{jin2023time} to perform this alignment. The patch reprogrammer performs cross attention between patches and a set of (transformed) token embeddings from the LLM. The result is a representation of each patch as a linear combination of pretrained token embeddings. This is a powerful inductive bias that improves the ability of the encoder to construct embeddings which match the expected inputs to the LLM.

While \citet{jin2023time} develop a time series forecasting method using the patch reprogrammer, their method treats each covariate independently, so no information mixing between covariates can occur. While this may be suitable for some applications, there are many applications in medicine for which interaction effects are very relevant. For example, ventilator waveform anomalies can accompany changes in vital signs such as increased heart rate or decreased oxygen saturation and increase a clinician's concern for clinically significant dyssynchrony. We therefore introduce a set of methods for incorporating covariate features into the model. There are many possible ways of incorporating these features, each with complex tradeoffs. We propose four approaches: concatenating the covariate signals for each patch, averaging the patch embeddings across covariates, interleaving the embeddings, and averaging the final predictions.

In a multivariate setting, we have a signal $x_{1:T,1:C} \in \mathbb{R}^{T \times C}$ with patches $\{ x_{t:t+l-1,c} : c \in \mathbb{N}_{\leq C}, i \in P, t = i s + 1 \}$.
The `concatenation' strategy passes each covariate signal through the normalization layer and initial embedding layer independently, and concatenates the patch embeddings before the reprogramming layer, forming combined patch embeddings $\phi_t \in \mathbb{R}^{d_p \times C} \forall t \in P$, where $d_p$ is the dimension of each univariate patch embedding.
The `averaging' strategy keeps the embeddings of each covariate independent until after the patch reprogrammer, at which point it computes a weighted average $\phi^\prime_t = \sum_{c \in \mathbb{N}_{\leq C}} w_c \phi_{t,c}$ using learned weights $w_c \in \mathbb{R}_{\geq 0}$ such that $\sum_c w_c = 1$.
The `interleave' strategy, instead of merging the embeddings, interleaves them in the input to the LLM, constructing a sequence $(\phi_{1,1},...,\phi_{1,C},\phi_{2,1},...,\phi_{2,C},...,\phi_{N,1},...,\phi_{N,C})$. The information from each covariate is therefore mixed in the LLM, and combined in a final projection layer of the model.
Finally, the `independent' strategy treats each covariate independently throughout the model, and averages the final outputs.
Visualizations of these methods are provided in \cref{fig:model}, and a full comparison in \cref{sec:discussion}.

\subsection{LLM Backbone}
\label{sec:methods:llm}

The core component of our model that enables it to leverage unstructured text data is a pretrained LLM. Rather than training our own domain-specific LLM, we make use of state-of-the-art publicly available pretrained foundation models. Previous studies have demonstrated that models trained on vast swaths of general data can significantly outperform specialized models \citep{Nori2023CanGF}, so we aim to harness their capabilities for the medical time series domain.
In order to preserve the language understanding abilities gained through pretraining and minimize computational requirements, we use the LLM in a frozen state, without fine-tuning its parameters.
We do not limit our framework to a particular LLM, instead it can be used with any current LLM architecture, as further explored in \cref{sec:appendix:results:llm-abl}.

The input to the frozen pretrained LLM is constructed by concatenating the token embeddings produced by the LLM's embedding layer for the textual prompt with the patch embeddings produced by the time series encoder. This allows the model to utilize the semantic information from both the time series and the unstructured metadata in its predictions.
After passing the inputs through the model, we extract the token embeddings produced by the final layer. 

Since decoder-only transformers are sequence-to-sequence models, the LLM backbone computes an output token for each input token, including both text tokens and time series patches. While the LLM uses the text tokens to generate outputs for each time series patches, the model’s purpose is to make predictions only for the time series patches, allowing the outputs corresponding to the text tokens to be safely ignored. The output embeddings corresponding to the time series patches are concatenated and the resulting vector is fed into a linear projection layer, which maps the concatenated embedding vector directly to the output time series points, to make the final predictions for the selected time series analysis problem.

\subsection{Task Solvers}
\label{sec:methods:tasks}

\subsubsection{Semantic Segmentation}
\label{sec:methods:task:semseg}

Semantic segmentation is the task of partitioning a time series into discrete segments and assigning a classification to each segment. This can be used to identify each phase of an event (such as inspiratory and expiratory phases of a breath), or pick out specific types of events (such as different types of heartbeat arrhythmias).
While this task can be solved through a process of boundary detection then segment classification, these two sub-tasks are inter-related -- in particular, the classification of a segment can influence the optimal boundary selection. Therefore, we instead predict a class for each point in the input time series. Point-wise classifications can then be assembled into classified segments by grouping contiguous sets of points with the same predicted label. This approach solves the problem in a single pass, combining the segmentation and classification sub-tasks.

To predict the label for each point, we first predict the likelihood of each class for each point.
In the projection layer following the LLM, we map the set of final patch embeddings from the LLM to a matrix of size $N \times C$, where $N$ is the number of time points, and $C$ is the number of classes. We then apply the {\sc softmax} function to each row, resulting in a class likelihood vector for each point. We use cross-entropy loss for training, and take the {\sc argmax} over the likelihood vectors to compute the predicted label for each point during testing.
In settings where there are only two classes, it is not necessary to predict the likelihood of both as they must sum to 1. Therefore, we predict only a single value for each point, and use the {\sc sigmoid} function to map it to the likelihood of one class. The likelihood is then thresholded at 0.5 to determine the predicted class for every point. In this case, the model is trained with the binary cross entropy loss function. 

The Semantic Segmentation pipeline in \cref{fig:model} shows how patch embeddings are projected to multiple class values per time point. After applying the {\sc softmax} function, we compute the class label for each point, and at inference, consecutive points with the same class label are combined to form segments.

\subsubsection{Boundary Detection}
\label{sec:methods:tasks:segmentation}

Boundary detection involves partitioning a time series into discrete segments by predicting the boundaries of each segment.
It differs from semantic segmentation primarily in that segments are not assigned classifications, and each event that is segmented may be semantically identical to its neighbors. 
While there is less information to predict in boundary detection, it is typically more challenging than semantic segmentation because it is difficult to delineate between consecutive periods that may be very similar.
MedTsLLM performs boundary detection by classifying each boundary point.

In our boundary point classification method, we treat boundary detection as a binary classification problem, in which every point is either a boundary point between segments or not. For each point in the time series, we predict a single value representing the likelihood of the point being a segment boundary. The model is trained using binary cross entropy (BCE) loss.
While the labels are extremely unbalanced, which often poses significant challenges for classifiers, we have found that this method performs well in practice, which we attribute to our approach for identifying boundaries from the raw likelihood scores.
Rather than threshold these scores at 0.5, or some other fixed level, we apply an algorithm that finds local maxima \citep{2020SciPy-NMeth} of the score signal, which correspond to the most likely boundary points. We enforce a constraint that no pair of selected points is closer than some distance threshold parameter, which ensures that only one point is selected for each region of elevated scores.
An approximately optimal distance parameter can be found using gradient-free optimization methods that maximize the target evaluation metric over the training data, or validation data if available. We also utilize the 10$^\text{th}$ percentile of segment lengths in the training set as a simple but effective heuristic that is much faster to compute. \Cref{fig:model}'s Boundary Detection section illustrates how after projecting patch embeddings to points representing likelihood scores, the model is trained using BCE to compare the scores against the true binary class labels. At the inference stage, using the described constrained peak-finding algorithm, we identify the most likely peaks in the likelihood scores that ultimately correspond to boundary points.

\subsubsection{Anomaly Detection}
\label{sec:methods:tasks:anom}

The final time series analysis task we perform is unsupervised anomaly detection. In this setting we assume we have an unlabeled training set of ``normal'' periods, and we aim to detect anomalous periods that deviate from normal in unseen data.
We utilize the standard approach to this task in the machine learning literature, which consists of training a model to reconstruct the normal input signals, and during inference, marking points in the input as anomalous if they deviate from the predicted signal by more than some threshold.
This can be an effective approach as the reconstruction model is only trained to predict normal sequences, so significant deviations from its predictions are likely to be anomalies.
More specifically, we train our model to take an input signal and output a signal as similar as possible by minimizing the mean square error (MSE). During testing, the model attempts to reconstruct an input signal, and we compute an anomaly score for each point, which is the MSE between the inputs and predictions, normalized across features. We set a threshold using the frequency of anomalies in each dataset and the distribution of scores, and predict that any points with score above this threshold are anomalous. The Anomaly Detection section in \cref{fig:model} illustrates how the MSE between the denormalized, predicted (reconstructed) signal and the original time series signal is thresholded to identify periods of anomalous points.

\section{Datasets}
\label{sec:datasets}
We showcase our model's applicability on multiple datasets spanning two domains: ECGs and respiratory signals. The ventilator data specifically is internally collected with patient consent and approved by the relevant institutional review board. We also discuss the processing of publicly available datasets, extracted from Physionet \citep{goldberger2000physiobank}. Additional summary statistics can be found in \cref{tab:semantic-seg-stats,tab:boundary-detection-stats,tab:anom-detect-stats}.

\vspace{-.3cm}
\begin{table}[ht]
    \centering \footnotesize
    {\renewcommand{\arraystretch}{1.15}
    \caption{\small Overview of the datasets used in this study. These statistics pertain to our processed versions of the datasets that we use for analysis rather than the raw data. Only semantic segmentation datasets have classifications.}
    \begin{tabular}{p{2cm}|llllll}
        \textbf{Dataset} & \textbf{Category} & \textbf{Task} & \textbf{Features} & \textbf{Classes} & \textbf{Time points} \\
        \hline
        Ventilator & Respiration & Semantic segmentation & 2 & 2 & 988,217 \\
        LUDB & ECG & Semantic segmentation & 1 & 4 & 8,771,395 \\
        BIDMC & Respiration & Boundary detection & 3 & -- & 3,180,053 \\
        MIT-BIH & ECG & Boundary detection & 2 & -- & 9,027,800 \\
        MIT-BIH Arrhythmia & \multirow{2}{*}{ECG} & \multirow{2}{*}{Anomaly detection} & \multirow{2}{*}{2} & \multirow{2}{*}{--} & \multirow{2}{*}{7,447,935}%
    \end{tabular}}
    
    \label{tab:dataset-stats}
    \vspace{-0.5cm}
\end{table}

\subsection{Ventilator Waveforms}
\label{sec:datasets:ventilator}

To study mechanical ventilation for pediatric patients, we have, with IRB approval, internally collected a dataset of ventilator waveforms and relevant clinical information from N=17 patients from the Pediatric Intensive Care Unit at Johns Hopkins All Children's Hospital from July 2020 to August 2021. The dataset consists of over 1,700 hours of EHR data (e.g. patient demographics, medication administration, etc.) and physiologic time series data (e.g. numeric values and waveforms from GE physiologic monitors and Draeger ventilators).
10 30-minute clips of ventilator pressure and flow waveforms, extracted from a curated subset of 5 patients on pressure-control synchronized intermittent mandatory ventilation, were segmented into inspiratory and expiratory periods. An expert clinician annotated 60\% of the waveforms using ECG lead II as a reference, while a trained intern annotated the remaining 40\%, which was then checked by the supervising clinician. Small segments that could not be cleanly segmented following expert guidelines were removed, resulting in a total of 7,344 identified breaths.

The dataset includes 7 clips from periods with a ventilator-measured triggered rate of at most 1 (stable, ventilator-delivered breaths) and 3 clips with rates between 2 and 5 (patient-triggered, often noisier breaths), providing a physiologically diverse dataset representative of different patient states on ventilatory support. To facilitate waveform analysis, the prompt part of the dataset includes clip-specific information (e.g. statistics on both ventilator-derived signals like respiratory rate and other signals like heart rate, and ventilator settings), and patient-specific information (e.g. age, gender, medications, etc.).

\subsection{Publicly Available Datasets}
\paragraph{Lobachevsky University Electrocardiography Database (LUDB). }
\label{sec:datasets:ludb}

LUDB, a public ECG delineation dataset \citep{kalyakulina2020ludb, kalyakulina2021lobachevsky}, contains 200 10-second-long 12-lead ECG signals sampled at 500 Hz from 200 healthy volunteers and patients with various cardiovascular diseases. Two cardiologists annotated each lead with P wave, T wave, and QRS complex boundaries and peaks for each beat, along with patient diagnoses. The dataset also includes patient information such as sex, heart rhythms, and conduction abnormalities.
Classifying every point into one of the three classes (P, T, and QRS), or a fourth ``unlabeled'' class, is a semantic segmentation task. Each ECG lead is considered a separate univariate time series due to independent annotation. 80\% of patients were randomly selected for the training set, and the rest were used for testing.

\paragraph{BIDMC PPG and Respiration}
\label{sec:datasets:bidmc}

The BIDMC PPG and Respiration dataset \citep{7748483}, collected from critically-ill patients as part of the MIMIC II matched waveform database \citep{lee2011open}, consists of 53 8-minute recordings of various physiological signals (sampled at 125 Hz) and physiological parameters (sampled at 1 Hz). The analysis focuses on signals consistently available across all patients: the impedance respiratory signal, plethysmograph, and ECG lead II. Averages of each physiological parameter such as heart rate are computed for context, along with patient age and sex.
Two annotators segmented the data into individual breaths using the impedance respiratory signal, with labels consisting of boundary points between consecutive breaths. For simplicity, only the first annotator's labels are utilized.
The dataset is partitioned into training (85\% of patients) and test (15\% of patients) sets.

\paragraph{MIT-BIH}
\label{sec:datasets:mitbih}

The MIT-BIH dataset \citep{moody2001impact} consists of 48 30-minute, 2-channel ambulatory ECG recordings from 47 patients. For consistency, we include only patients with MLII and V1 ECG channels and classify beat annotations as either normal or abnormal, with any label other than normal (e.g., left bundle branch block beat) considered abnormal.
The dataset also contains patient information, including age, medications, and annotator notes about identified artifacts in the waveforms.
Following \citet{ecg_sampling_freq}, we downsample the signals from 360Hz to 125Hz to reduce the computational demands without compromising relevant information.

We solve two tasks on this dataset: boundary detection for segmenting beats and anomaly detection for identifying arrhythmias, with different splitting and label processing strategies that reflect the particulars of each task.
For boundary prediction, labels correspond to the annotated beat peak points, with no distinction between normal and abnormal beats.
Patients are partitioned randomly between training (80\%) and testing (20\%) subsets.
For anomaly detection, we select patients with the least number of anomalies for training, and those with the most anomalies for testing, maintaining an 80/20 split. Patients with all abnormal ECG beats were excluded. 
Each abnormal annotation is expanded to fill the window of 150 ms before and after the annotation, following the American Medical Instruments standard \citep{aami1999nsi}.

\paragraph{Non-Medical Datasets}
\label{sec:datasets:nonclinical}

As we only have one medical time series dataset for anomaly detection, to demonstrate our method's ability to perform across datasets, we include two additional non-medical datasets frequently used for benchmarking time series anomaly detection methods: the Mars Science Laboratory (MSL) dataset, a public NASA dataset containing expert-labeled telemetry anomaly data \citep{DBLP:journals/corr/abs-1802-04431}, and %
the Pooled Server Metrics (PSM) dataset, which contains internally collected data from multiple application server nodes at eBay \citep{abdulaal2021practical}.

\section{Results}
\label{sec:results}

In this section, we demonstrate the potential of MedTsLLM to solve semantic segmentation, boundary detection, and anomaly detection on our selected datasets (see \Cref{tab:dataset-stats}), and evaluate its performance relative to other state-of-the-art methods based on metrics and experimental setup defined in \cref{sec:results:eval-method}. %
We evaluate our model on various task-solvers in \crefrange{sec:results:semseg}{sec:results:anom}, and perform ablation studies on prompting and covariate strategies in \cref{sec:results:ablation}.

\subsection{Evaluation Approach} 
\label{sec:results:eval-method}

\paragraph{Metrics.}
We utilize task-specific metrics to evaluate the performance of our framework.
For semantic segmentation, we use the mean segment-wise Intersection over Union (mIoU) and point-wise F1 score, which are commonly used to assess the quality of segmentation predictions in various domains. 
For anomaly detection, we report the F1 score and Area Under the Receiver Operating Characteristic curve (AUROC), as they are standard metrics for binary classification problems that provide a balanced measure of performance. We compute these metrics period-wise, as opposed to point-wise, using standard procedure employed in benchmarking recent unsupervised anomaly detection methods \citep{wu2022timesnet, Xu_2018}. 
To evaluate boundary detection performance, we report the segment-wise mIoU and the accuracy of predicted segments with at least 0.75 IoU overlap with ground truth segments. We also measure the point-wise mean absolute error (MAE) between predicted and actual boundary points. While these are not standard metrics for time series boundary detection, we believe that the segment-wise mIoU is most relevant to downstream applications, while point-wise MAE provides a more granular measure of boundary localization performance.

\paragraph{Baseline methods.} 
We primarily compare the performance of our model against a variety of competitive models for time series analysis tasks that fall into three categories: LLM-based models, other general deep learning models, and domain-specific methods. We conduct a separate comparison with traditional time series analysis methods described in \cref{tab:trad-ts-sem-seg,tab:trad-ts-bound-det,tab:trad-ts-anom-det} in the appendix. 
The LLM-based approach, introduced by \citet{zhou2024one} (GPT4TS), fine-tunes selected layers of GPT-2 on time series patches, and has achieved competitive performance on benchmark datasets.
We select three deep learning models for general time series analysis: PatchTST \citep{nie2022}, TimesNet \citep{wu2022timesnet}, and FEDformer \citep{zhou2022fedformer}.
These three models were selected because they were consistently top performers in recent works that evaluate deep learning models on anomaly detection across a range of benchmark datasets.
While these LLM and deep learning methods can be used for anomaly detection without modification, none of them natively support semantic segmentation or boundary prediction. We therefore adapt our approach to these tasks for each model.
Additionally, we include two domain-specific methods that were developed to solve particular medical time series semantic segmentation tasks.
Ventiliser \citep{chong2021} is an algorithm designed to segment and classify breathing phases based on ventilator pressure and flow signals. We utilize this method as a baseline for semantic segmentation on our ventilator dataset.
On the LUDB dataset, we compare against \citet{utime2019}, which develops a U-Net inspired convolutional neural network to segment ECGs into onsets and offsets of the P and T waves and QRS complexes.

\paragraph{Experiment details.} 
Throughout \crefrange{sec:results:semseg}{sec:results:anom}, we employ a standardized approach across models for evaluation. Each model is trained for 10 epochs using identical training and task-solver parameters.
In these experiments, MedTsLLM uses LLama 2 (7B) \citep{touvron2023llama} as the backbone LLM, with dataset and task prompts, and the ``concatenate'' covariate strategy.
Further details and information about the implementation and code can be found in \cref{sec:appendix:setup}.

\subsection{Semantic Segmentation}
\label{sec:results:semseg}
\Cref{tab:results:semseg} presents the results of a semantic segmentation task performed by different models on the Ventilator and LUDB datasets. MedTsLLM achieves the highest F1 scores and IOU values on both datasets, indicating its effectiveness for semantic segmentation. While all models except the domain-specific model perform exceptionally well on the Ventilator dataset, the LUDB dataset appears to be more challenging, with a wider range of performance across the models. This suggests that deep learning models, particularly MedTsLLM and PatchTST, better capture the complex patterns and features present in the LUDB dataset.

\begin{table}[ht]
\vspace{-.5cm}
  \centering\small
  \caption{Semantic segmentation results.}
  \begin{tabular}{lllll}
    \toprule
    \multirow{2}{*}{Model} & \multicolumn{2}{c}{Ventilator} & \multicolumn{2}{c}{LUDB}\\
    \cmidrule(lr){2-3} \cmidrule(lr){4-5} 
    & F1 & IoU & F1 & IoU \\
    \hline
    MedTsLLM & \textbf{98.92} & \textbf{97.86} & \textbf{89.89} & \textbf{81.73} \\
    GPT4TS & 98.81 & 97.65 & 78.92 & 65.78 \\
    TimesNet & 98.55 & 97.14 & 76.24 & 62.44 \\
    PatchTST & 98.72 & 97.46 & 89.54 & 81.31 \\
    FEDformer & 98.66 & 97.35 & 74.97 & 62.62 \\
    Domain-specifc & 89.18 & 80.47 & 40.12 & 33.59 \\
    \bottomrule
  \end{tabular}
  \label{tab:results:semseg} 
\vspace{-.5cm}
\end{table}

\subsection{Boundary Detection}
\label{sec:results:seg}

\Cref{tab:results:seg} presents the results of boundary prediction performed by different models on two datasets: BIDMC and MIT-BIH. MedTsLLM emerges as the best-performing model for boundary detection on both datasets, demonstrating its strong segmentation capabilities across different domains. PatchTST and TimesNet also show strong performance, consistently ranking among the top three models, while FEDformer performs well on the BIDMC dataset, ranking second, but falls behind TimesNet on the MIT-BIH dataset. GPT4TS has the lowest performance on both datasets, suggesting that it may not be as well-suited for these specific segmentation tasks, despite being an LLM-based method.
\begin{table}[ht]
\vspace{-.5cm}
  \centering\small\footnotesize
  \caption{BIDMC and MIT-BIH boundary detection results.} \small
    \begin{tabular}{llrrrr}
    \toprule
        Datasets & Model &  mIoU &  \shortstack{Accuracy\\@ 0.75 IoU}  &  \shortstack{Change Point\\MAE} &  \shortstack{Accuracy\\(~50 pts)} \\
    \midrule
         BIDMC & MedTsLLM & \textbf{0.87} &               \textbf{0.84} &           \textbf{32.59} &                       \textbf{0.84} \\
        &  GPT4TS & 0.64 &               0.29 &           95.13 &                       0.32 \\
        &   TimesNet & 0.71 &               0.47 &           72.94 &                       0.53 \\
        &    PatchTST & 0.75 &               0.69 &           65.24 &                       0.59 \\
        & FEDformer & 0.85 &               0.82 &           32.95 &                       \textbf{0.84} \\
         \midrule
        MIT-BIH & MedTsLLM & \textbf{0.89} &                \textbf{0.90} &             \textbf{6.40} &                        \textbf{0.98} \\
        &  GPT4TS & 0.64 &                0.34 &            25.86 &                        0.82 \\
        &  TimesNet & 0.86 &                0.84 &             7.19 &                        0.97 \\
        &  PatchTST & 0.87 &                0.87 &             7.90 &                        0.97 \\
        & FEDformer & 0.84 &                0.77 &            10.01 &                        \textbf{0.98} \\

    \bottomrule
    \end{tabular}
    \label{tab:results:seg} 
\vspace{-.5cm}
\end{table}

\subsection{Anomaly Detection}
\label{sec:results:anom}
The results in \Cref{tab:results:anom} suggest that MedTsLLM's consistent top performance across all three datasets indicates its robustness and superior ability to detect anomalies in different types of time series data. The MIT-BIH dataset proves to be the most challenging, with models exhibiting varying levels of performance. The mixed results of GPT4TS and FEDformer across datasets indicate that their architectures or training approaches may not be as well-suited for anomaly detection tasks compared to MedTsLLM, PatchTST, and TimesNet.

\begin{table}[ht]
\vspace{-.5cm}
  \centering\small
  \caption{Anomaly detection results.}\small
  \begin{tabular}{llllllll}
    \toprule
    \multirow{2}{*}{Model} & \multicolumn{2}{c}{PSM} & \multicolumn{2}{c}{MSL}  & \multicolumn{2}{c}{MIT-BIH} \\
    \cmidrule(lr){2-3} \cmidrule(lr){4-5} \cmidrule(lr){6-7}
    & F1 & AUROC & F1 & AUROC & F1 & AUROC \\
    \midrule
    MedTsLLM & \textbf{97.31} & \textbf{98.20} & \textbf{88.00} & \textbf{90.95} & \textbf{94.70} & \textbf{98.52} \\
    GPT4TS & 90.23 & 91.40 & 72.51 & 82.29 & 72.19 & 80.83 \\
    TimesNet & 89.69 & 90.92 & 81.80 & 87.13 & 88.29 & 92.20 \\
    PatchTST & 95.12 & 95.60 & 78.65 & 84.97 & 89.53 & 93.70 \\
    FEDformer & 90.04 & 90.94 & 82.22 & 87.15 & 51.23 & 67.21 \\
    \bottomrule
  \end{tabular}
  \label{tab:results:anom} 
\vspace{-.5cm}
\end{table}

\subsection{Ablation Studies}
\label{sec:results:ablation}

We perform ablation studies to demonstrate the effects of changing the (1) covariates and (2) prompting strategies on model performance. Table \ref{tab:results:abl:covs} shows the performance of using different covariates on two tasks: segmentation (BIDMC) and anomaly detection (MIT-BIH). Results indicate that across both datasets and tasks, interleaving or concatenating covariates leads to the best performance.
Table \ref{tab:results:abl:prompt} shows how different ways of handling the prompt affect performance. The strategies with the word `only' in the label refer to only keeping that component in the prompt. Results show that using patient-specific information to contextualize the time series leads to the best gains in performance. We explore the implications of these results in the Discussion section below.

\begin{table}[htpb]
\vspace{-.5cm}
\centering\small
\caption{Results of our ablation study on covariate strategy.}
\label{tab:results:abl:covs}
\begin{tabular}{llllll}
\toprule
\multirow{2}{*}{Strategy} & \multicolumn{2}{c}{BIDMC} & \multicolumn{2}{c}{MIT-BIH (Anomalies)} \\
\cmidrule(lr){2-3} \cmidrule(lr){4-5}
& IoU & MAE & F1 & AUROC \\
\hline
Concatenate          & \textbf{86.00} & \textbf{37.53} & 94.46             & 96.75             \\
Interleave           & 84.40          & 42.91          & \textbf{95.75}    & \textbf{97.69}    \\
Average (weighted)   & 83.96          & 42.69          & 92.65             & 95.87             \\
Average (unweighted) & 83.74          & 43.88          & 82.25             & 90.68             \\
Independent          & 83.99          & 43.52          & 88.36             & 93.92             \\
\bottomrule
\end{tabular}
\vspace{-.5cm}
\end{table}
\begin{table}[htpb]
\centering\small
\caption{Results of our ablation study on prompting strategies.}
\label{tab:results:abl:prompt}
\begin{tabular}{lllll}
\toprule
\multirow{2}{*}{Strategy} & \multicolumn{2}{c}{LUDB} & \multicolumn{2}{c}{MIT-BIH (Boundaries)} \\
\cmidrule(lr){2-3} \cmidrule(lr){4-5}
& IoU & F1 & IoU & MAE \\
\midrule
No prompt    & 77.90          & 87.40          & 55.61                   & 32.51                  \\
Dataset only & 80.71          & 89.21          & 60.31                   & 22.10                  \\
Task only    & 80.02          & 88.79          & 92.17                   & 4.31           \\
Patient only & \textbf{80.82} & \textbf{89.31} & \textbf{92.18}          & \textbf{4.20}                   \\
Stats only   & 80.75          & 89.23          & 56.05                   & 31.11                  \\
All          & 79.50          & 88.41          & 57.07                   & 28.76                  \\
\bottomrule
\end{tabular}
\vspace{-.5cm}
\end{table}

\section{Discussion}
\label{sec:discussion}

MedTsLLM consistently outperforms state-of-the-art baselines, showcasing the potential of leveraging LLMs for analyzing complex physiological signals. In this section, we explore the technical and clinical implications of our methodology.

\subsection{Technical Implications}
\textbf{Domain-agnostic approach.}
Our study shows that our model can perform well across different medical applications, including respiration and ECG, and for different time series tasks. %
This provides evidence of the generalizability of LLM-based frameworks and suggests that LLMs may be useful for a diverse range of tasks in healthcare. In addition, using LLMs opens up opportunities for future work to make use of currently underutilized sources of rich medical data, like EHR, which contain semi-structured, heterogeneous data.

\noindent \textbf{Covariate strategies.}
Integrating information across interrelated signals is critical in multivariate time series analysis, particularly in the medical domain, where multiple physiologic signals can have complex interactions that should drive predictions.
Several recent deep learning methods for general time series forecasting have challenged this outlook \citep{nie2022, jin2023time}, including TimeLLM, which we derive our time series encoder from, and have found success on standard benchmark datasets by treating covariates independently.
However, our ablation studies convincingly demonstrate that across a range of medical tasks, our method performs significantly better when utilizing multivariate information.
Furthermore, covariate handling strategies that allow the model to best integrate information across covariates, in particular the interleave strategy, also perform best.

\Cref{tab:pros-cons-cov-strategies} summarizes the pros and cons of each strategy.
While task performance is ultimately the most consequential consideration in selecting a strategy, other factors may influence the optimal method for a particular dataset or task.
In particular, interleaving covariate tokens in the LLM input allows for flexibility in handling missing information, different frequencies, or irregular time steps, and allows the LLM to attend to each covariate patch and determine how to mix information across them.
However, it multiplies the size of the input sequence to the LLM by the number of covariates included, so it does not scale well to higher dimensional datasets. Instead, the embedding averaging strategy can be used which requires limited additional memory or computational power with increasing covariates, and retains the flexibility, but does not perform as well. A more balanced approach is concatenating covariate patches, which scales well and achieves strong performance.

\newcommand{\threestar}{{\color{Green3}$\bigstar\bigstar\bigstar$}}
\newcommand{\twostar}{{\color{Gold1}$\bigstar\bigstar$}}
\newcommand{\onestar}{{\color{red}$\bigstar$}}
\begin{table}[htpb]
\vspace{-.5cm}
\small
    \centering
    \caption{Comparision of different covariate strategies.}
    \begin{tabular}{lllll}
    \toprule
        \textbf{Property} & \textbf{Average} & \textbf{Concatenate} & \textbf{Interleave} &  \textbf{Independent} \\
    \midrule
    Memory scaling & \threestar & \twostar & \onestar & \twostar \\
    Information retention  & \onestar & \twostar & \threestar &\twostar \\
    Data flexibility& \threestar & \onestar & \threestar & \twostar \\
    Performance & \twostar & \threestar & \threestar & \onestar \\
    \bottomrule
    \end{tabular}
    \label{tab:pros-cons-cov-strategies} 
\vspace{-.3cm}
\end{table}

\noindent\textbf{Prompting strategy.}
Recent publicly available LLMs have been trained on massive a corpus of information collected from the internet, giving them ``knowledge'' of medical conditions, time series analysis, specific medical time series tasks, and other valuable information for our problem \citep{Singhal2022LargeLM}.
Effective prompting of the LLM is essential to ensure that it can utilize this knowledge and apply it to its predictions. We therefore have constructed prompting strategies to provide the necessary context and instructions to maximize the performance of LLMs on our tasks.
Information about the general domain of the problem and the specific dataset are included to ground the LLM's predictions in this critical context.
Patient-specific information allows for incorporating additional unstructured data for each time series, which is not otherwise accessible for time series models.
Statistics about the input signals and other low frequency data in text format help to provide information about the signals in the modality that LLMs are trained to use.
Finally, a task instruction tells the LLM what its outputs should be, which can be particularly valuable if using instruction or chat-tuned LLMs.
Our prompting ablation studies demonstrate the value of each of these prompt components in improving the performance of MedTsLLM.
However, we find that with too many of these prompt components used at once, performance can suffer.
We hypothesize that this is likely due to the LLM struggling to determine what information is important when the prompt grows too large, which has been observed in other settings \citep{liu2023lost}, and is a critical problem to address in future work.

\subsection{Clinical Implications}

Analyzing physiological signals is crucial for clinical decision-making, as insights extracted from them help clinicians better understand patient state in real time. Our three tasks enable different types of analysis that contribute to this understanding. Using ECGs as an example, our model's accurate boundary detection enables heart rate calculation and heart rate variability analysis, providing critical insights into cardiovascular health. Semantic segmentation further delineates heartbeats into P and T waves and the QRS complex, which are essential for diagnosing cardiac disease states. Accurate and timely anomaly detection allows clinicians to monitor a patient's state and make responsive treatment decisions.

\textbf{Potential in dyssynchrony analysis.}
We further explore the applicability of our model within the context of mechanical ventilation. Mechanical ventilation can be life-threatening when suboptimal settings result in patient-ventilator dyssynchrony (PVD), which clinicians struggle to detect and manage due to information overload and limited time for observation before needing to intervene. Addressing this challenge requires automated and accurate detection of PVD. Current dyssynchrony research necessitates precise breath segmentation labels obtained from pressure and flow waveforms. Unfortunately, the availability of publicly accessible, labeled, high-frequency ventilator waveform datasets and breath segmentation tools is severely limited compared to other medical domains, such as ECG analysis. 
    
Our work offers valuable contributions in two significant ways. Firstly, with just 5 hours of labeling, our model performs better than clinical segmentation tools, without even enforcing specific breath waveform types. This could encourage other research teams to utilize our tool for building more breath segmentation datasets. Secondly, our model's anomaly detection abilities could be harnessed for PVD detection. Although we could not validate its performance due to the lack of anomaly labels, the potential is promising. One common obstacle faced by researchers in the dyssynchrony space is the extensive resources required for segmenting data into inspiration and expiration phases before labeling PVD. Our model has the potential to streamline this process by aiding in both segmentation and identifying anomalies or dyssynchronies for clinical review.

\textbf{Leveraging underutilized multimodal data.}
One of the key factors that contribute to our model's success is its ability to incorporate heterogeneous data modalities, which is particularly useful when dealing with a mixture of time series signals and EHRs. EHRs contain data of various modes, frequencies, temporal resolutions, and distributions, which traditional time series models cannot handle effectively (and hence do not include). In contrast, our natural language prompting strategy can fuse these diverse data types without requiring conversion to standardized, structured format. This enables inclusion of the wide-ranging data extracted from each patient, and our ablation studies \ref{tab:results:abl:prompt} confirm that it does significantly improves performance. By incorporating raw time series signals, our model offers an alternative to existing LLMs that process EHR data solely through clinical notes or require inputs to be in a specific format. This flexibility enables our model to leverage the full potential of the diverse data available from each patient, leading to more accurate and comprehensive analyses.

\subsection{Limitations and future directions}
Despite our promising method and strong results presented in our work, there are several limitations that should be addressed for future research. Firstly, while MedTsLLM demonstrates high performance, interpretability of the model's predictions remains a challenge. Future work should focus on developing methods to explain the model's decisions and provide more transparent insights to clinicians, perhaps through text output. To note, however, our focus is on relatively straightforward, lower-level tasks that are too time-intensive for humans to perform at scale, not decision-making. The high volume of model predictions renders it impractical for clinicians to audit each one. In these situations, we believe model performance is more critical for trustworthiness than interpretability is, and have hence prioritized the former. 

Secondly, compared with simpler time series models, our model is more computationally intensive to train. In this study, we have decided to focus on maximizing our method's performance rather than its computational efficiency, especially as ML methods become increasingly prevalent in medicine, and as clinical environments update their infrastructure accordingly. Nonetheless, while training LLM models of various sizes, we discovered that using smaller ones can significantly reduce computational requirements with only a small drop in performance (\Cref{tab:llm-abl}). Future work can explore optimizations for LLM inference like quantization and key-value (KV) caching, to be used in resource-constrained environments.

Our current approach involves freezing the LLM backbone and training specific layers. While fine-tuning the LLM backbone on domain-specific medical data could potentially enhance its adaptation to healthcare applications, the computational demands of our method render this impractical. Even with parameter-efficient techniques like LoRA \citep{hu2021lora}, the increased cost and complexity are likely to outweigh any performance benefits.  

It is important to note that our method does require fine-tuning specific layers for each dataset and task. Future research could explore training an LLM from scratch to inherently comprehend time series data, which might also improve generalizability. However, this approach faces a significant challenge, one we encountered when seeking to expand our method to other medical domains: the scarcity of suitable public clinical datasets, especially outside cardiology and pulmonology. The limited availability of large-scale, appropriately labeled time series datasets in other medical domains has constrained the scope of such endeavors.

One alternate way to further realize the potential of MedTsLLM in real-world clinical settings is incorporating more EHR data and extensive clinical notes to help the LLM build a more detailed profile of each patient. Another future direction would be adding more task functionality to MedTsLLM, such as forecasting, clustering, and classification. In combination, these directions can contribute to the development of more powerful, transparent, and widely applicable tools for clinical decision support and personalized medicine.

\subsection{Conclusion}

Our work introduces MedTsLLM, a novel approach that leverages the power of large language models for medical time series analysis. By integrating patient-specific contextual information and handling multiple covariates, MedTsLLM outperforms state-of-the-art baselines on critical tasks such as boundary detection, semantic segmentation and anomaly detection. This work represents a significant step towards building general-purpose models that effectively combine insights extracted from multimodal data and knowledge from pretrained LLMs, paving the way for more accurate, actionable, and personalized insights from multi-dimensional physiological signals. Clinically, MedTsLLM has the potential to transform patient monitoring, clinical decision support, and personalized medicine, ultimately improving patient outcomes and advancing the field of healthcare.

\newpage
\bibliography{medtsllm-arxiv}

\newpage

\acks{This work was carried out at the Advanced Research Computing at Hopkins (ARCH) core facility  (rockfish.jhu.edu), which is supported by the National Science Foundation (NSF) grant number OAC1920103.}

\appendix

\section{Additional dataset information}

\begin{table}[ht]
    \centering 
    \caption{Descriptions of each dataset that are used for prompting MedTsLLM.}
    \begin{tabular}{l|p{0.85\textwidth}}
        \toprule
        \textbf{Dataset} & \textbf{Description} \\
        \midrule
        Ventilator & The dataset contains time series data of airway pressure and flow rate measurements collected from a mechanical ventilator during the respiratory support of a fully sedated patient. The data is sampled at a frequency of 100 Hz. The airway pressure is measured in cmH$_2$O and the flow rate is measured in L/min. \\
        \hline
        LUDB & LUDB is an ECG signal database collected from subjects with various cardiovascular diseases used for ECG delineation. Cardiologists manually annotated boundaries of P, T waves and QRS complexes. Each clip consists of a 10-second signal from a single ECG lead, sampled at 500 Hz. \\
        \hline
        BIDMC & The BIDMC dataset is a dataset of electrocardiogram (ECG), pulse oximetry, photoplethysmogram (PPG) and impedance pneumography respiratory signals acquired from intensive care patients. Two annotators manually annotated individual breaths in each recording using the impedance respiratory signal. \\
        \hline
        MIT-BIH & The MIT-BIH Arrhythmia Database contains excerpts of two-channel ambulatory ECG from a mixed population of inpatients and outpatients, digitized at 360 samples per second per channel with 11-bit resolution over a 10 mV range. \\
    \end{tabular}
    \label{tab:dataset-prompts}
\end{table}

\begin{table}[ht]
  \centering 
  \caption{Semantic segmentation dataset statistics.}
  \small
  \begin{tabular}{@{}llllllll@{}}
  \toprule
    \textbf{Dataset} & \textbf{\shortstack{Health\\domain}} & \textbf{\shortstack{\#\\dimensions}} & \textbf{\shortstack{\#\\classes}} & \textbf{\shortstack{\#\\Train}} & \textbf{\shortstack{\#\\Test}} & \textbf{\shortstack{\# Train class\\distribution}} & \textbf{\shortstack{\# Test class\\distribution}} \\
    \midrule
    Ventilator & Respiration & 3 & 2 & 395,665 & 69,823 & \begin{tabular}[t]{@{}p{2cm}@{}}I: 0.267\\E: 0.709\end{tabular} & \begin{tabular}[t]{@{}p{2cm}@{}}I: 0.229\\E: 0.771\end{tabular}\\ 
    LUDB & ECG & \begin{tabular}[t]{@{}p{1.7cm}@{}}1 (12 leads fed independently)\end{tabular} & 4 & 6,988,275 & 1,783,120 & \begin{tabular}[t]{@{}p{2cm}@{}}P: 0.091 N: 0.122\\T: 0.199 U: 0.589\end{tabular} & \begin{tabular}[t]{@{}p{2cm}@{}}P: 0.103 N: 0.127\\T: 0.127 U: 0.565\end{tabular} \\ 
    \bottomrule
  \end{tabular}
  \label{tab:semantic-seg-stats} 
\end{table}

\begin{table}[htbp]
  \centering
  \caption{Boundary detection dataset statistics.}
  \begin{tabular}{@{}llllllll@{}}
    \toprule
    \textbf{Dataset} & \textbf{\shortstack{Health\\domain}} & \textbf{\shortstack{\#\\dimensions}} & \textbf{\shortstack{\#\\classes}} & \textbf{\shortstack{\#\\Train}} & \textbf{\shortstack{\#\\Test}} & \textbf{\shortstack{Train\\boundary\\point\\ratio}} & \textbf{\shortstack{Test\\boundary\\point\\ratio}} \\
    \midrule
    BIDMC & Respiration & 3 & 2 & 2,520,042 & 660,011 & 0.011389 & 0.00946 \\
    MIT-BIH & ECG & 2 & 2 & 1,128,475 & 315,973 & 0.002296 & 0.002251 \\
    \bottomrule
  \end{tabular}
  \label{tab:boundary-detection-stats}
\end{table}

\begin{table}[ht]
  \centering 
  \caption{Anomaly detection dataset statistics.}
  \small
  \begin{tabular}{@{}llllll@{}}
  \toprule
    \textbf{Dataset} & \textbf{Application} & \textbf{\# dimensions} & \textbf{\# Training} & \textbf{\# Test} & \textbf{Anomaly ratio} \\
    \midrule
    MSL & Space & 55 & 58,317 & 73,729 & 0.105 \\ 
    PSM & Server & 26 & 132,481 & 87,841 & 0.278 \\
    MIT-BIH Arrhythmia & Health & 2 & 1,354,170 & 902,780 & 0.261 \\ 
    \bottomrule
  \end{tabular}
  \label{tab:anom-detect-stats} 
\end{table}

\clearpage
\section{Additional Results}
\label{sec:appendix:results}

\subsection{LLM Ablation Study}
\label{sec:appendix:results:llm-abl}

\begin{table}[ht]
\centering
\caption{Ablation study results comparing performance across LLM backbones for semantic segmentation on the LUDB dataset.}
\label{tab:llm-abl}
\begin{tabular}{l|ll}
\toprule
\textbf{LLM}             & \textbf{Parameters} & \textbf{IoU}   \\
\midrule
Llama 2 7b Chat & 6.7B       & 81.73 \\
BioMedLM        & 2.7B       & 81.25 \\
Mamba 2.8b      & 2.8B       & 81.09 \\
GPT2 XL         & 1.6B       & 80.50 \\
Llama 2 7b      & 6.7B       & 80.25 \\
Mamba 1.4b      & 1.4B       & 79.76 \\
GPT2            & 137M       & 78.11 \\
\bottomrule
\end{tabular}
\end{table}

\subsection{Traditional time series results comparison}
\label{sec:appendix:results:trad-ts-comparison}

\begin{table}[htbp]
    \centering 
    \caption{\small Comparing semantic segmentation results for MedTsLLM with traditional time series analysis methods on the LUDB dataset.}
    \begin{tabular}{p{2cm}|llll}
        \toprule
        \textbf{Method} & \textbf{IoU} & \textbf{F1} & \textbf{Description}  \\
        \hline
        MedTsLLM & \textbf{97.86} & \textbf{98.92} & Our proposed method \\
        Thresholding & 89.30 & 94.34 & Predict expiration for points with flow<0.05 \\
        KNN & 91.47 & 95.54 & Point-wise K-nearest neighbors classifier \\
HMM & 90.48 & 95.00 & Hidden Markov Model \\
    \bottomrule
    \end{tabular}
    \label{tab:trad-ts-sem-seg}
\end{table}

\begin{table}[htbp]
\centering 
\caption{\small Comparing boundary detection results for MedTsLLM with traditional time series analysis methods on the BIDMC dataset.}
\begin{tabular}{p{3.5cm}|llp{8cm}}
    \toprule
    \textbf{Method} & \textbf{mIoU} & \textbf{MAE} & \textbf{Description} \\
    \midrule
    MedTsLLM & \textbf{86.56} & \textbf{32.59} & Our proposed method \\
    Peak Detection & 74.96 & 94.67 & Segment boundaries are usually at peaks of the RESP signal, so run SciPy's peak detection algorithm. \\
    Template Matching & 70.82 & 81.369 & Select 20 template breaths and use dynamic time warping distance to find segments. \\
    \bottomrule
\end{tabular}
\label{tab:trad-ts-bound-det}
\end{table}

\begin{table}[htbp]
\centering
\caption{\small Comparing anomaly detection results for MedTsLLM with traditional time series analysis methods on the MIT-BIH dataset.}
\begin{tabular}{p{2.5cm}|llp{8cm}}
    \toprule
    \textbf{Method} & \textbf{F1} & \textbf{AUROC} & \textbf{Description} \\
    \midrule
    MedTsLLM & \textbf{94.7} & \textbf{98.52} & Our proposed method \\
    Quantile-based & 81.11 & 88.23 & Flag points outside specified quantile thresholds (e.g., 5\textsuperscript{th} and 95\textsuperscript{th} percentiles). \\
    Z-score & 72.52 & 82.23 & Flag points beyond a set number of standard deviations (z-score) from the mean. \\
    Rolling average & 55 & 69.18 & Flag points deviating from the rolling average beyond a set threshold. \\
    FFT-based & 52.58 & 68.02 & Threshold reconstruction error from an FFT-based model. \\
    \bottomrule
\end{tabular}
\label{tab:trad-ts-anom-det}
\end{table}

\section{Experimental Setup and Implementation}
\label{sec:appendix:setup}

Our models were implemented using PyTorch (version 2.2.1) \citep{ansel2024} and the Transformers library (version 4.39) \citep{wolf2020}. Training used the Ranger optimizer \citep{wright2021} with an initial learning rate of 0.0001 for 15 epochs. Training was performed on NVIDIA A100 80G GPUs.
Code for this project can be found at: \href{https://github.com/flixpar/med-ts-llm}{https://github.com/flixpar/med-ts-llm}.

\end{document}